\documentclass{article}

\PassOptionsToPackage{numbers,compress}{natbib}
\usepackage[dblblindworkshop, final]{neurips_2026}
\workshoptitle{Long-Context Foundation Models}

\usepackage[utf8]{inputenc}
\usepackage[T1]{fontenc}
\usepackage[hidelinks]{hyperref}
\usepackage{url}
\usepackage{booktabs}
\usepackage{amsfonts}
\usepackage{amsmath,amssymb}
\usepackage{nicefrac}
\usepackage{microtype}
\usepackage{xcolor}
\usepackage{graphicx}
\usepackage{multirow}
\usepackage{array}
\usepackage{placeins}

\newcolumntype{L}[1]{>{\raggedright\arraybackslash}p{#1}}
\newcolumntype{C}[1]{>{\centering\arraybackslash}p{#1}}
\graphicspath{{figures/}{media/}{img/}{./}}


\title{Measurement-First Auditing of Agentic Leaderboards:
Contamination Susceptibility, Matched-Control Re-evaluation, and
Scorer Validation}

\author{%
  Dishu Yang \\
  Khoury College of Computer Sciences \\
  Northeastern University \\
  San Jose, CA, USA \\
  \texttt{yang.dis@northeastern.edu} \\
  \And
  Qi Su \\
  School of Engineering and Applied Science \\
  The George Washington University \\
  Bellevue, WA, USA \\
  \texttt{qisu623@gwu.edu} \\
  \AND
  Hongbo Qin \\
  Khoury College of Computer Sciences \\
  Northeastern University \\
  Boston, MA, USA \\
  \texttt{qin.ho@northeastern.edu} \\
  \And
  Hansong Zhang \\
  Department of Computer Science \\
  University of California, Berkeley \\
  Berkeley, CA, USA \\
  \texttt{hansong@berkeley.edu} \\
}

\begin{document}
\raggedbottom

\maketitle

\begin{abstract}
Agentic leaderboards increasingly evaluate systems on public benchmarks
whose task statements and solution-bearing artifacts can remain
accessible. We propose a measurement-first audit framework that
distinguishes contamination claims according to the evidence required to
support them. It separates three channels that require different evidence: training-time exposure, evaluation-time retrieval, and pipeline/scaffold leakage. Each channel is coded as open, partial, closed, or unknown under a fail-closed rule. Across nine Holistic Agent Leaderboard (HAL) configurations, none of the 27 channel assessments was coded closed, but incidents were confirmed in four configurations. We then apply the behavioral
component of the framework to a reported file-localization gap on
SWE-bench Verified, using an outcome-blind, same-repository
matched-control design with symmetric prompt-leakage screening, paired
and repository-aware uncertainty analyses, and scorer validation,
evaluated on GPT-4.1 and DeepSeek-V4-Flash. Among the 100 pairs retained
after symmetric screening and the pair-integrity exclusion, GPT-4.1
showed a $+10.0$-point pair-weighted Top-3 benchmark-associated gap, but the 95\% intervals from both the prespecified paired-bootstrap
procedure and the post-hoc repository-balanced analysis included zero,
leaving the benchmark-associated gap inconclusive. The reproduction scorer did not pass its validation gate: against
consensus human labels, sufficient scorer sensitivity could not be
established for either model, and both DeepSeek-V4-Flash firings on
correct-gold comparisons were false positives. Without provenance
evidence, appropriate controls, symmetric leakage screening, and
validated scorers, stronger contamination claims are not warranted. The
results do not establish training-data membership, contamination
prevalence, or benchmark-induced score inflation.
\end{abstract}

\section{Introduction}
\label{sec:intro}

Public coding benchmarks are often treated as evidence that an agent can repair real bugs, yet when task statements and solution-bearing artifacts remain publicly accessible, benchmark performance may partly reflect recall rather than repair. 

The Holistic Agent Leaderboard (HAL) evaluates nine benchmark configurations spanning coding, web navigation, science, and customer service~\cite{kapoor2026hal}. A contamination claim about such a score usually conflates three questions that require different evidence. Were solution-bearing artifacts available for training? Could the evaluated configuration reach them at run time? Did the harness itself place gold information in prompts or evaluator inputs? Behavioral gaps are also open to multiple explanations. Liang \emph{et al.}~\cite{liang2026illusion} report markedly higher file-path accuracy on SWE-bench Verified than outside the benchmark ecosystem, without an uncertainty interval, yet repository composition, issue length, or patch complexity could contribute to such a gap. SWE-bench Verified is a human-filtered subset selected for well-specified issues~\cite{openai2024verified}, so the gap may reflect issue quality rather than memorization.

Our goal is therefore not to establish that SWE-bench is contaminated, but to provide a framework for determining which contamination claims the available evidence can and cannot support. The framework comprises four components: (i) a revision-pinned provenance audit, (ii) an outcome-blind, same-repository matched-control design, (iii) symmetric answer-leakage screening, and (iv) validation of the reproduction scorer. Stage I audits the three channels across nine HAL configurations using revision-pinned evidence and a fail-closed coding rule. Stage II re-evaluates two diagnostic probes from Liang \emph{et al.}~\cite{liang2026illusion} against comparable public controls, with symmetric answer-leakage screening and repository-aware uncertainty analyses. For the reproduction probe, scorer false-positive behavior and sensitivity are assessed against consensus human labels to determine whether its outputs support behavioral inference.

Applied to HAL and SWE-bench Verified, the framework finds broad susceptibility but fewer confirmed incidents. File-localization gaps are positive but inconclusive, while scorer validation leaves the reproduction claims unresolved.

\section{Related work}
\label{sec:related}

The configurations we audit run within HAL~\cite{kapoor2026hal}.
Search-equipped agents can surface benchmark ground truth on public
mirrors at evaluation time~\cite{han2025searchtime}. Our susceptibility card (Section~\ref{sec:threat}) records
channel prerequisites, supporting evidence, and unknowns
for each configuration and revision.

The most directly related work, Liang \emph{et al.}~\cite{liang2026illusion}, introduces file-path identification and function reproduction. They report higher file-path accuracy on SWE-bench Verified than
on outside-repository tasks, and score reproduction by 5-gram overlap. We adapt these probes using same-repository matching, symmetric leakage screening, explicit uncertainty analysis, and scorer validation for false-positive behavior and sensitivity. 

Contamination detection includes corpus-side overlap~\cite{riddell2024quantifying}, behavioral memorization probes~\cite{prathifkumar2025memory}, and paired confidence tests~\cite{zhang2024pacost}. Related privacy-auditing work treats canary-based memorization and PREMIA-based membership-inference signals as empirical diagnostics rather than formal privacy guarantees~\cite{yang2026balancing}. In benchmark audits, such signals alone do not identify how benchmark information became available; our framework therefore separates the three compromise channels, while Stage II estimates a benchmark-associated contrast rather than a membership probability. We use standard matching methods~\cite{stuart2010matching} and cluster-aware uncertainty for shared repositories~\cite{cameron2015cluster}. Because overlap metrics can reward surface similarity rather than the intended construct~\cite{xiao2023evaluating}, we assess Probe 2 scorer validity before drawing behavioral conclusions.

\section{Methodology}
\label{sec:method}

\subsection{Threat framework and inference scope}
\label{sec:threat}

We consider three non-mutually exclusive compromise channels. \textbf{Training-time exposure} concerns whether task statements and solution-bearing artifacts were publicly available, task-linkable, and temporally eligible relative to a model. \textbf{Evaluation-time retrieval} concerns whether the assessed
configuration provided a retrieval mechanism, a prompt-visible lookup
key, and positive evidence that the relevant external or local-corpus
retrieval route was not technically blocked. These conditions establish
an opportunity to retrieve benchmark artifacts, not evidence that
retrieval occurred. \textbf{Pipeline/scaffold leakage} concerns whether tasks, gold artifacts, hidden-test information, or reference outputs entered prompts, examples, scaffolds, or evaluator inputs at a documented revision.

We record these assessments in a structured susceptibility and incident card, where each row represents a benchmark-level audit entry with its in-scope HAL configurations, channel labels, and linked evidence at a pinned revision. Training-time exposure is assessed at benchmark scope unless model-specific temporal evidence is available, whereas retrieval and pipeline findings are configuration- and revision-specific. Each channel is coded open (O), partial (P), closed (C), or unknown (?). O requires all predefined observable prerequisites or a confirmed revision-specific incident. P requires affirmative or partial support for at least one prerequisite without establishing the full pathway. C requires complete negative evidence within scope. ? indicates no affirmative or partial support and insufficient evidence for closure. Missing evidence is not treated as negative evidence. Confirmed incidents are recorded separately: an incident establishes a realized pathway at a specific revision, whereas an O label denotes susceptibility. We neither combine channel labels into an aggregate score nor interpret them as proof of contamination. The resulting susceptibility and incident card is reported in Table~\ref{tab:hal_card}.

\subsection{HAL-wide susceptibility and incident audit}
\label{sec:hal-audit}

We applied the card to nine HAL configurations using a fixed HAL
snapshot and versioned registry. The primary evidence search was
frozen on August 25, 2026. Revision-pinned checks during coding and
adjudication extended through September 9. Automated procedures
collected metadata and flagged candidate evidence, but final judgments
were human-coded: two authors independently coded 117 field-level
judgments, agreeing on 90 (76.9\%) before adjudication (Appendix~\ref{app:b}). Final T/R/L labels were derived mechanically from the adjudicated fields.

Three artifact-based audits supplemented the provenance review:
a local inventory of task artifacts, prompts, retrieval mechanisms,
and evaluator boundaries; a 40-query search-discoverability audit
over 20 SWE-bench Verified Mini tasks; and a comparison of locally
available benchmark-test and non-test artifacts, with trace scanning
where execution traces were available. Search hits, overlap matches,
and trace flags were treated only as leads; confirmation required
author review of the source, benchmark identity, threat channel, and
revision. Appendix~\ref{app:c} gives the procedures.

\subsection{Matched-control behavioral audit}
\label{sec:matched}

We test whether benchmark-associated behavioral
differences persist after matching each SWE-bench Verified task
one-to-one, without replacement, to a same-repository merged bug-fix
pull request not included in SWE-bench. Controls are matched on task
timing, issue size, and pull-request complexity under fixed
common-support constraints. Cohort construction, matching, and leakage
filtering are completed without access to model responses or probe
outcomes.

Building on Liang \emph{et al.}~\cite{liang2026illusion}, we adapt two probes. Benchmark and control arms use identical prompt templates, eligibility rules, and scoring procedures. 

\textbf{Probe 1: file-path identification.} Given the repository name and a sanitized issue description, the model predicts up to three production-source files modified by the accepted fix. Predictions are scored against production-source paths in the complete accepted pull request, with Top-3 accuracy as the prespecified primary endpoint.

\textbf{Probe 2: conditional solution reproduction.} Given the repository name, sanitized issue description, and the path of one modified production-source file, the model reconstructs the key code change. The target construct is key-line recovery from the accepted patch. We use a stricter exact-line scorer in a post-hoc, validation-gated analysis; scorer construction and secondary diagnostics are reported in Appendix~\ref{app:a}. 

Leakage filtering is symmetric at the matched-pair level. Probe 1 excludes a pair if either issue reveals a complete gold path or basename. Probe 2 excludes a pair if, after whitespace normalization, either
issue contains a non-comment production-code line added by the
accepted patch with at least five lexical tokens and 20 non-whitespace
characters. Matches found only in doctest or REPL examples are
recorded separately and do not trigger exclusion. Both arms are removed whenever either is directly revealing. A separate post-hoc pair-integrity audit removes any pair whose benchmark and control PRs address the same underlying issue; the single identified collision is excluded from both probes without replacement. Probe 1 inference therefore applies to pairs passing both the leakage and pair-integrity screens rather than the original matched cohort.

Probe 2 scorer validity is assessed in two directions. First, strict-scorable responses are rescored against deterministically selected, disjoint wrong-gold signatures from other tasks in the same repository to assess false-positive behavior on constructed negatives. Second, two authors independently review a stratified enriched sample, blinded to benchmark-control membership but not model identity or scorer-derived stratum. Consensus labels support design-weighted sensitivity estimation and assessment of correct-gold scorer firings. Sampling, weighting, label definitions, and validation-metric definitions are reported in Appendix~\ref{app:a}. Validation gates the use of Probe 2 contrasts to adjudicate H2 and H3; the strict-score analyses remain post hoc and exploratory.

\subsection{Statistical analysis and interpretation}
\label{sec:stats}

For a fixed metric, let $I$ denote the retained matched pairs, with a
common complete-case restriction across models where needed, and let $Y(\cdot) \in [0,1]$ denote the probe outcome. For pair $i$, let $b_i$ and $c_i$ denote the benchmark and control tasks. The prespecified pair-weighted
estimand is
\begin{equation}
\Delta_{\mathrm{pair}}
= \frac{1}{|I|}\sum_{i\in I}[Y(b_i)-Y(c_i)].
\label{eq:pair}
\end{equation}
For repository $r$, let $\Delta_r$ be the mean paired difference
among its retained pairs. The post-hoc repository-balanced estimand is
\begin{equation}
\Delta_{\mathrm{repo}}
= \frac{1}{R}\sum_r \Delta_r,
\label{eq:repo}
\end{equation}
where $R$ is the number of represented repositories. $\Delta_{\mathrm{pair}}$ averages over pairs, whereas $\Delta_{\mathrm{repo}}$ weights represented repositories equally.

Uncertainty for the prespecified pair-weighted estimand uses a paired
percentile bootstrap with 10{,}000 replicates, resampling matched pairs
and not explicitly modeling repository-level dependence. Post hoc, we
added the repository-balanced estimand with a $t$-interval over
repository-specific gaps ($R-1$ degrees of freedom; $R\geq3$), together
with hierarchical repository-and-pair bootstrap and leave-one-repository-out
sensitivity analyses. Because the pair-weighted and repository-balanced
analyses differ in both estimand and interval construction, differences
between them cannot be attributed solely to either component. Analyses
are conducted separately by model and metric; intervals are pointwise
and unadjusted, so secondary interval exclusions are treated as nominal
and no family-wise confirmatory claim is made.

Probe 2 contrasts are used to adjudicate H2 and H3 only if the wrong-gold
audit supports low false-positive behavior on constructed negatives,
correct-gold scorer firings are confirmed by consensus human labels,
and design-weighted review supports sufficient sensitivity. The outcome-blind amendment specified no numerical threshold for adequate scorer sensitivity.
The gate requires affirmative evidence of sufficient sensitivity;
inconclusive validation does not count as a pass. If the gate is not
passed, Probe 2 contrasts remain descriptive and H2 and H3 unresolved.

These estimands characterize benchmark-associated differences within
the common-support matched sample; they do not establish training-data
membership, a causal effect of benchmark exposure, or benchmark-induced
score inflation. ``Prespecified'' refers only to the original analysis
plan; Appendix~\ref{app:d} distinguishes original,
outcome-blind-amended, and post-hoc choices.

\section{Experiments and results}
\label{sec:results}

\subsection{Experimental setup}
\label{sec:setup}

\textbf{Cohort construction and matching.} We constructed an outcome-blind matched-control cohort from a pinned revision of the 500-task SWE-bench Verified test split~\cite{jimenez2024swebench}.
We retained all 50 HAL Mini tasks and sampled 150 additional tasks using
deterministic repository-stratified sampling with seed 13. Both benchmark
statements and control issues were required to contain 30 to 1,200 words,
and repository quotas were capped at 30 tasks and fixed before control
retrieval.

Controls were drawn from 5,172 merged, issue-linked PRs passing a frozen lexical bug-fix screen after excluding all pinned SWE-bench Full, Lite, and
Verified instances. Feasible pairs came from the same repository, had
merge dates within 365 days, differed by at most one changed file and
200 issue words, and shared the same temporal category relative to the
documented GPT-4.1 cutoff of June 1, 2024~\cite{openai2025gpt41}.
All retained benchmark and control PRs fell before this boundary. No
documented cutoff was available for DeepSeek-V4-Flash, so these
categories do not establish temporal eligibility for that model. The
public controls are not confirmed unexposed negatives for either model.
Minimum-cost one-to-one matching was performed separately within
repositories without control reuse (Appendix~\ref{app:d}).

One benchmark task lacked a feasible control, leaving 199 matched pairs
(50 HAL-linked and 149 expanded). A post-hoc pair-integrity audit removed
one same-issue collision from both probe analyses without replacement.
The original 199 pairs remain the denominator for the prompt-leakage
audit. Post-match balance on ten covariates satisfied
$|\mathrm{SMD}|<0.20$ in the pooled, HAL-linked, and expanded cohorts
(Appendix~\ref{app:d}). Probe-specific leakage filtering then followed
Section~\ref{sec:matched}.

\textbf{Models and inference settings.}
We evaluated GPT-4.1 and DeepSeek-V4-Flash with temperature 0, one
completion per task, and no tools, browsing, or retrieval. Output
budgets, retry handling, and additional inference settings are reported
in Appendix~\ref{app:d}. The matched pair is the unit of analysis.
Estimands and interval procedures follow Section~\ref{sec:stats}.

\begin{figure}[htbp]
\centering
\includegraphics[width=0.7\linewidth]{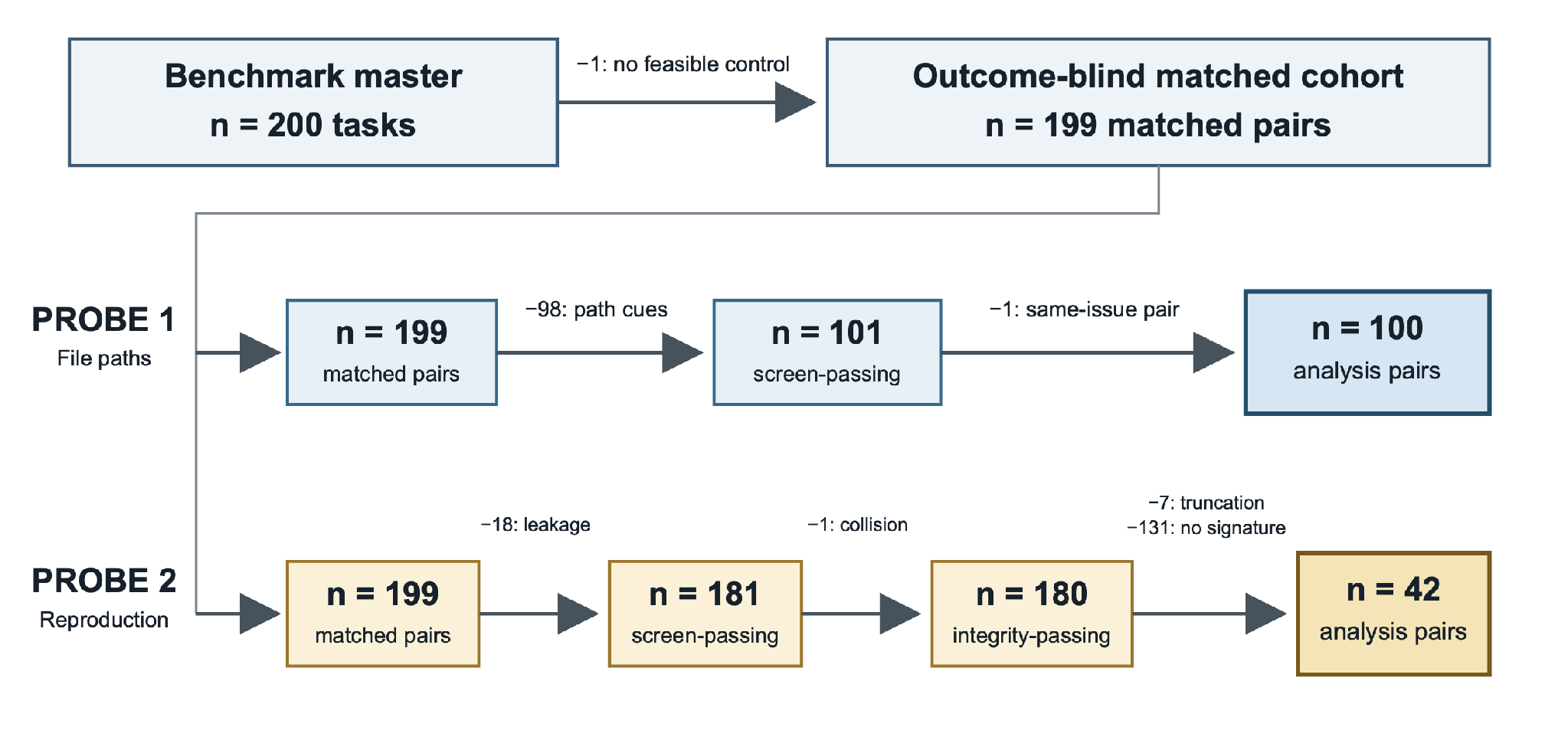}
\caption{Stage II cohort construction and probe-specific attrition. All
post-matching counts refer to matched pairs. Leakage screening excluded
both members when either arm met the exclusion criterion. The same-issue
exclusion removed one pair whose two arms addressed the same underlying
issue. For Probe 2, the final arrow combines sequential exclusions for
persistent truncation (7 pairs) and absence of an eligible strict
signature in at least one arm (131 pairs).}
\label{fig:cohort}
\end{figure}

\textbf{Analysis samples and hypotheses.} Probe 1 retained 100 matched pairs across ten repositories after
symmetric leakage screening and the pair-integrity exclusion
(Figure~\ref{fig:cohort}).  All ten balance covariates remained below 0.20 (maximum absolute SMD: 0.116).
Probe 2 retained 42 strict-scorable pairs across nine repositories; a
32-pair task-type-concordant subset was used for sensitivity analysis
(Appendix~\ref{app:a}).

We prespecified three hypotheses: H1 predicted higher file-path accuracy for benchmark tasks than matched controls; H2 predicted higher patch-specific line recovery; and H3 predicted that reproduction would be rarer than localization and, conditional on adequate scorer sensitivity, would show a benchmark-control gap at least as large as localization.

\subsection{HAL-wide susceptibility and incident findings}
\label{sec:hal-findings}

Across the nine HAL configurations, training-time exposure was partial
for all nine, evaluation-time retrieval was open for seven and partial
for two, and pipeline/scaffold leakage was open for four, partial for
two, and unknown for three (Table~\ref{tab:hal_card}). All training-time
labels were partial because temporal eligibility, which is
model-scoped, was unknown for every configuration; eight of the nine
had affirmative support for the other three prerequisites, and Online
Mind2Web additionally lacked public gold and task-gold linkability, so
the uniform P reflects the aggregation rule rather than homogeneous
evidence. No channel was rated closed: under the fail-closed rule, missing traces or comparison artifacts remain unknown rather than evidence of isolation, so these labels indicate susceptibility rather than contamination probability.

\begin{table}[t]
\caption{HAL-wide susceptibility and incident card. No channel was
coded closed; confirmed incidents are narrower than susceptibility.}
\label{tab:hal_card}
\centering
\footnotesize
\begin{tabular}{@{}L{5.3cm} l c c L{2.1cm} l@{}}
\toprule
& \multicolumn{3}{c}{\textbf{Susceptibility}}
& \multicolumn{2}{c}{\textbf{Realized incident}} \\
\cmidrule(lr){2-4}\cmidrule(lr){5-6}
\textbf{Benchmark} & \textbf{T} & \textbf{R} & \textbf{L}
& \textbf{Confirmed} & \textbf{Revision} \\
\midrule
Online Mind2Web (BrowserUse / SeeAct)      & P [Y/N/N/?] & O & ? & Not established & Unknown \\
AssistantBench (Browser agent)             & P [Y/Y/Y/?] & O & O & L               & Current \\
GAIA (Open Deep Research)                  & P [Y/Y/Y/?] & O & ? & Not established & Unknown \\
CORE-Bench Hard (CORE-Agent / Generalist)  & P [Y/Y/Y/?] & O & P & Not established & Unknown \\
ScienceAgentBench (SAB Self-Debug)         & P [Y/Y/Y/?] & P & P & None            & ---     \\
SciCode (Tool-calling agent)               & P [Y/Y/Y/?] & O & O & L               & Current \\
SWE-bench Verified Mini (SWE-Agent / Generalist) & P [Y/Y/Y/?] & O & ? & Not established & Unknown \\
USACO (Episodic + Semantic)                & P [Y/Y/Y/?] & O & O & L               & Historical \\
TAU-bench Airline (Few-shot / Generalist)  & P [Y/Y/Y/?] & P & O & L               & Current \\
\bottomrule
\end{tabular}

\vspace{3pt}
\begin{minipage}{\linewidth}
\footnotesize\raggedright
\textit{Notes.} T, R, L: training-time exposure, evaluation-time
retrieval, pipeline/scaffold leakage; O, P, C, ?: open, partial, closed,
unknown (Section~\ref{sec:threat}). ``Not established'' means the
evidence could neither confirm nor rule out a realized incident;
``None'' is a bounded negative judgment within the audited scope.
Bracketed T entries give, in order, public task availability, public
solution-bearing artifacts, task-artifact linkability, and temporal
eligibility (Y yes, Pt partial, N no, ? unknown).
\end{minipage}
\end{table}

Confirmed incidents were substantially narrower than channel-level
susceptibility. Four configurations had confirmed pipeline incidents
(current in AssistantBench, SciCode, and TAU-bench Airline; historical
in USACO), and no configuration-scoped retrieval incident was
established. The HAL repair record documents 29 GAIA mirror visits for
Generalist runs, but the audited GAIA row covers Open Deep Research, so
we leave that judgment unknown rather than transfer an incident across
scaffolds. ScienceAgentBench received a bounded none judgment; Online
Mind2Web, GAIA, CORE-Bench Hard, and SWE-bench Verified Mini remained
unknown.

In the focused SWE-bench discoverability demonstration (20 tasks, 40
frozen queries, one web-search service, August 25, 2026), 35 of 40
queries returned a task-linked result in the top ten; gold information
was directly reachable for 7 queries, partially reachable for 23, and
not reachable for 10. These counts describe discoverability under one
service on one date, not a prevalence estimate, and do not show that an
evaluated agent retrieved the artifacts or that they appeared in
training data. We selected SWE-bench Verified for the matched-control
analysis because it supported task-level provenance and same-repository
control construction, not because it had the most severe susceptibility
profile.

\subsection{Probe 1: file-path identification}
\label{sec:probe1}

After symmetric prompt-leakage filtering and exclusion of the cross-arm
underlying-issue collision, Probe 1 analyzed 100 matched pairs across
ten repositories. Within this analysis subset, both pair-weighted and
repository-balanced benchmark-associated gap estimates were positive for
all eight model-metric combinations (Table~\ref{tab:probe1}). On the
prespecified primary endpoint, Top-3 accuracy was 0.640 versus 0.570 for
DeepSeek-V4-Flash and 0.710 versus 0.610 for GPT-4.1. The corresponding
pair-weighted gaps were $+0.070$ (95\% CI $[-0.070, +0.210]$) and
$+0.100$ ($[-0.030, +0.230]$); the post-hoc repository-balanced gaps
were $+0.081$ ($[-0.079, +0.240]$) and $+0.109$ ($[-0.116, +0.334]$).
Because pairs are clustered within ten repositories, we also used a
hierarchical bootstrap that resampled repositories and then pairs
within repositories; the resulting Top-3 95\% intervals were $[-0.132, +0.283]$ for
DeepSeek-V4-Flash and $[-0.128, +0.334]$ for GPT-4.1. Thus, none of the
pair-weighted, repository-balanced, or hierarchical analyses yielded a
Top-3 interval excluding zero.

For GPT-4.1, three secondary-metric intervals nominally excluded zero
under the prespecified pair-weighted analysis, whereas none did under
the post-hoc repository-balanced analysis. All DeepSeek-V4-Flash
intervals included zero under both analyses. Within the analysis subset, GPT-4.1 and DeepSeek-V4-Flash showed Top-3 gaps of $+10.0$ and $+7.0$ points, respectively, directionally
consistent with Liang \emph{et al.}~\cite{liang2026illusion}; the
different cohorts and metrics preclude a magnitude comparison.

\begin{table}[!htbp]
\caption{Probe 1 benchmark-associated gaps (benchmark minus control),
100-pair subset. All gaps are positive; no Top-3 interval excludes
zero.}
\label{tab:probe1}
\centering
\footnotesize
\begin{tabular}{@{}l l C{4.3cm} C{4.3cm}@{}}
\toprule
\textbf{Model} & \textbf{Metric}
& \textbf{Pair-weighted gap [95\% CI]}
& \textbf{Repository-balanced gap [95\% CI]} \\
\midrule
\multirow{4}{*}{DeepSeek-V4-Flash}
 & Top-3 (primary) & $+0.070$ $[-0.070, +0.210]$ & $+0.081$ $[-0.079, +0.240]$ \\
 & Top-1           & $+0.100$ $[-0.030, +0.230]$ & $+0.086$ $[-0.081, +0.252]$ \\
 & MRR@3           & $+0.088$ $[-0.042, +0.217]$ & $+0.081$ $[-0.072, +0.233]$ \\
 & Path F1         & $+0.084$ $[-0.040, +0.205]$ & $+0.078$ $[-0.057, +0.213]$ \\
\cmidrule(lr){1-4}
\multirow{4}{*}{GPT-4.1}
 & Top-3 (primary) & $+0.100$ $[-0.030, +0.230]$ & $+0.109$ $[-0.116, +0.334]$ \\
 & Top-1           & $+0.160$ $[+0.020, +0.300]$ & $+0.138$ $[-0.051, +0.328]$ \\
 & MRR@3           & $+0.130$ $[+0.005, +0.258]$ & $+0.120$ $[-0.078, +0.319]$ \\
 & Path F1         & $+0.119$ $[+0.017, +0.222]$ & $+0.131$ $[-0.053, +0.314]$ \\
\bottomrule
\end{tabular}

\vspace{3pt}
\begin{minipage}{\linewidth}
\footnotesize\raggedright
\textit{Notes.} Pair-weighted intervals: prespecified paired
percentile bootstrap, 10{,}000 replicates. Repository-balanced
intervals: post-hoc $t$-interval over ten repository-specific gaps
($df = 9$). Controls are public tasks, not confirmed unexposed
negatives, so gaps are benchmark-associated differences rather than
contamination effects.
\end{minipage}
\end{table}

\subsection{Prompt-side answer leakage}
\label{sec:leakage}

The symmetric leakage audit found at least one gold source path or
basename verbatim in 63 of 199 benchmark issues (31.7\%) and 57 of 199
control issues (28.6\%). Excluding both members of any pair in which
either arm leaked removed 98 pairs (49.2\%), leaving 101; the cross-arm
underlying-issue exclusion removed one more, leaving 100 pairs for
Probe 1. The similar arm-level rates indicate that explicit path cues
are common in issue-driven localization rather than a pronounced
benchmark-control asymmetry, and the 31.7\% benchmark-arm rate is
close to the 27.0\% reported by Liang \emph{et
al.}~\cite{liang2026illusion}. We therefore treat the audit as a
measurement control, not as new evidence of benchmark-specific
compromise. Pair-level exclusion rates varied from 26.7\% (Sphinx) to
76.0\% (scikit-learn), making unaudited results sensitive to repository
composition, and because detection is verbatim these rates are lower
bounds.

The screen also selects on issue characteristics: retained benchmark
issues were shorter (172 versus 328 words), more HAL-linked, less often
task-type concordant, and differently distributed across repositories
than excluded ones (Appendix~\ref{app:g}). Probe 1 estimates therefore
describe a shorter-issue subpopulation in which explicit path cues were
absent from both arms, and within-subset balance does not establish
that the subset represents the cohort.

\subsection{Probe 2: scorer validation}
\label{sec:probe2-validation}

Probe 2 was treated as a validation-gated measurement study rather than
a confirmatory test of H2 or H3. Only 42 of the 199 matched pairs were
strict-scorable, and the scorer fired on just 3 of 168 model-arm
outputs. Automatic benchmark-control contrasts and
secondary metrics are reported in Table~\ref{tab:a1} as descriptive
diagnostics only.

Validation against independently assigned human labels showed that the
strict scorer missed most human-positive responses
(Appendix~\ref{app:a}, Table~\ref{tab:probe2}). After dichotomizing ratings at the $\geq 2$
threshold, Cohen's $\kappa$ was 0.788 for DeepSeek-V4-Flash and 1.000
for GPT-4.1. In the enriched validation samples, it detected 0 of 2
consensus-positive DeepSeek-V4-Flash responses and 1 of 3
consensus-positive GPT-4.1 responses, corresponding to design-weighted
sensitivity estimates of 0.00 and 0.19. The corresponding unweighted
exact 95\% intervals were $[0.00, 0.84]$ and $[0.01, 0.91]$. Given only two DeepSeek-V4-Flash and three GPT-4.1 consensus human-positive responses, respectively, these near-unit-width intervals do not establish the scorer sensitivity required by the validation gate. Both DeepSeek-V4-Flash firings were human-negative, whereas the single GPT-4.1 firing was human-positive. The wrong-gold audit produced no matches, but this constructed-negative check does not establish general specificity. Because the scorer did not pass the validation gate, H2 and H3 remain unresolved. This is a measurement failure, not evidence that solution reproduction is rare. The enriched samples were constructed to
assess scorer validity, not to estimate the population prevalence of solution reproduction.

\subsection{Hypothesis disposition and robustness}
\label{sec:disposition}

H1 is inconclusive: all point estimates were positive, but no Top-3
primary interval excluded zero. H2 and H3 are unresolved because the
scorer did not pass the validation gate, so the strict-score contrasts
remain exploratory and the sensitivity condition H3 requires could not
be established (Appendix~\ref{app:e}, Table~\ref{tab:hyp}).

Leave-one-repository-out re-estimation shifted the four reported
repository-balanced gaps by at most 0.068. In the post-hoc
64-pair task-type-concordant analysis, all Top-3 intervals included
zero (Appendix~\ref{app:a}, Table~\ref{tab:a3}).

\section{Discussion}
\label{sec:discussion}

\subsection{Interpretation and implications}
\label{sec:interpretation}

Taken together, the results support a layered rather than binary interpretation of benchmark compromise. The HAL-wide audit identifies observable opportunities for training-time exposure, evaluation-time retrieval, and pipeline/scaffold leakage, whereas confirmed incidents establish narrower, revision- and procedure-specific failures. Susceptibility indicates possible compromise, whereas a confirmed incident shows that a specific pathway was realized. The Stage II behavioral contrast estimates a benchmark-associated difference without identifying training contamination as its cause.

File-localization gaps were positive but inconclusive. Unmeasured issue
clarity remains a plausible alternative explanation, while leakage
screening limits inference to a shorter-issue subset. Because the
reproduction scorer did not pass the validation gate, Probe 2 contrasts
remain descriptive and do not establish reproduction frequency.

These findings support a measurement-first approach to benchmark auditing, in which provenance and, where available, execution-trace evidence are considered alongside symmetric leakage controls, matched public controls, and scorer checks for false positives and sensitivity. Rather
than assigning a binary ``clean'' or ``contaminated'' label, the
framework ties claim strength to the evidence available for each audit
component. 

\subsection{Limitations and threats to validity}
\label{sec:limitations}

Three features limit Probe 2. First, the enriched human-review sample supports diagnostic scorer
assessment, but its sparse consensus-positive labels do not establish sufficient sensitivity or population-level reliability. Second, the probe discloses the gold file path and scores
only against the accepted patch, crediting solution-specific recall
given the edit location while penalizing alternative valid
implementations. Third, manual review was blinded to arm but not to
model identity or scorer-derived stratum, so raters knew which
responses the scorer had flagged; this visibility may bias judgments
toward agreement with the scorer. With the validation gate unmet,
Probe 2 contrasts remain descriptive.

The Probe 2 sample is also selected by outcome-adjacent criteria and
fails the cohort-specific issue-length balance check. The strict
gold-signature requirement conditions on a property of the accepted
patch, while control candidates alone had to pass a lexical bug-fix
screen (Appendix~\ref{app:a}).

Statistical power, screening, and scope further bound the results.
Probe 1 inference is limited to the 100 pairs that remained after symmetric path-cue screening and the pair-integrity exclusion. Compared with the excluded pairs, this subset had shorter issues and a higher proportion of HAL-linked tasks. Probe 2 retained 42 pairs. Because these samples are clustered within ten and
nine repositories, respectively, intervals are wide, and intervals
including zero should not be read as evidence of no effect. The study
covers ten mature Python repositories and two API-served,
non-reasoning models, limiting generalization across languages,
projects, and model families. Controls are public tasks that predate the
documented GPT-4.1 cutoff rather than confirmed unexposed negatives, so
the design does not establish absence of pretraining exposure.
Generation used no fixed API seed, so committed completions cannot be
regenerated exactly.

The HAL audit is likewise specific to the benchmarks, scaffolds,
revisions, and evidence dates examined. Missing run traces and
configuration-attribution gaps limit several judgments, and
pre-adjudication agreement was lower for retrieval, leakage, and
incident judgments than for training-time judgments. Adjudication
resolves coding disagreements without removing those evidence limits.
Future work should evaluate semantically aware reproduction scorers on
larger, independently annotated samples and extend the matched-control
design across additional repositories and model families.

\section{Conclusion}
\label{sec:conclusion}

We presented a measurement-first audit framework that determines which
contamination claims the available evidence can support. Across nine
HAL configurations, no channel was coded closed and adjudicated
incidents were narrower than channel-level susceptibility. In the file-localization probe, the GPT-4.1 primary Top-3 gap was positive within a 100-pair, shorter-issue subset, but both its paired-bootstrap and repository-balanced 95\% intervals included zero. In the reproduction probe, the scorer did not pass the validation gate, so the instrument does not support the measurement. Neither outcome establishes training-data membership, a causal effect of benchmark exposure, or benchmark-induced score
inflation.

\begin{ack}
The authors received no external funding for this work and declare no
competing financial interests.
\end{ack}

\bibliographystyle{unsrtnat}

\appendix

\makeatletter
\@addtoreset{table}{section}
\makeatother
\renewcommand{\thetable}{\Alph{section}\arabic{table}}

\section{Probe 2 sample construction and descriptive diagnostics}
\label{app:a}

Benchmark identifiers, pull-request numbers, accepted patches, URLs, and explicit issue or commit identifiers were withheld or sanitized symmetrically. Probe 1 predictions were normalized before scoring; Top-1 accuracy, MRR@3, and exact-path set F1 were secondary metrics. For Probe 2, when multiple production-source files qualified, we deterministically selected the file with the most preliminarily eligible added lines before model execution. Example- and doctest-only overlaps were recorded separately in the Probe 2 leakage audit.

Strict gold signatures exclude blank lines, comments, imports,
declarations, and detected docstrings; retained lines require at least
eight lexical tokens and 30 characters, with at most 20 lines retained
per task. A response scores positive when it contains a
whitespace-normalized exact match to at least one signature line, and
tasks without eligible signatures are treated as measurement attrition.
Token F1 and gold-denominator 4-gram recall are reported as secondary
diagnostics. Because the strict scorer did not pass the validation gate,
all benchmark-control contrasts below are descriptive and do not
adjudicate H2 or H3.

\begin{table}[htbp]
\caption{Probe 2 sample construction and descriptive automatic results.}
\label{tab:a1}
\centering
\footnotesize

{\raggedright\textbf{Panel A. Sample construction}\par}
\vspace{3pt}
\begin{tabular}{@{}L{7.0cm} c c@{}}
\toprule
\textbf{Stage} & \textbf{Change} & \textbf{Pairs retained} \\
\midrule
Initial matched sample                        & ---   & 199 \\
Pair-level leakage exclusion                  & $-18$ & 181 \\
Cross-arm underlying-issue collision          & $-1$  & 180 \\
Persistent truncation                         & $-7$  & 173 \\
No eligible strict signature in benchmark arm & $-89$ & 84  \\
No eligible strict signature in control arm   & $-42$ & 42  \\
Task-type-concordant subset                   & ---   & 32  \\
\bottomrule
\end{tabular}

\vspace{10pt}
{\raggedright\textbf{Panel B. Repository-balanced automatic contrasts}\par}
\vspace{3pt}
\begin{tabular}{@{}l l l c C{2.0cm} C{2.3cm}@{}}
\toprule
\textbf{Model} & \textbf{Sample} & \textbf{Metric} & \textbf{N pairs}
& \textbf{Benchmark-control gap} & \textbf{95\% $t$-CI} \\
\midrule
\multirow{4}{*}{DeepSeek-V4-Flash}
 & Full strict          & Key-line recovery & 42 & $0.000$  & $[-0.077, 0.077]$ \\
 & Full strict          & Token F1          & 42 & $+0.018$ & $[-0.037, 0.074]$ \\
 & Full strict          & 4-gram recall     & 42 & $+0.020$ & $[-0.118, 0.158]$ \\
 & Task-type concordant & Key-line recovery & 32 & $+0.015$ & $[-0.090, 0.120]$ \\
\cmidrule(lr){1-6}
\multirow{4}{*}{GPT-4.1}
 & Full strict          & Key-line recovery & 42 & $-0.022$ & $[-0.073, 0.029]$ \\
 & Full strict          & Token F1          & 42 & $+0.009$ & $[-0.064, 0.082]$ \\
 & Full strict          & 4-gram recall     & 42 & $-0.021$ & $[-0.128, 0.087]$ \\
 & Task-type concordant & Key-line recovery & 32 & $-0.022$ & $[-0.073, 0.029]$ \\
\bottomrule
\end{tabular}

\vspace{3pt}
\begin{minipage}{\linewidth}
\footnotesize\raggedright
\textbf{Notes.} Panel A reports sequential exclusions. Among the 173
pairs surviving the leakage, collision, and truncation screens, strict
signatures were unavailable for 89 benchmark and 87 control arms when
assessed marginally. In the final 42-pair sample, issue-length SMDs were
$+0.152$ overall, $+0.241$ in the expanded cohort, and $-0.293$ in the
HAL-linked cohort. Panel B reports repository-balanced gaps with
$t$-intervals across nine repositories ($df = 8$). Because the scorer
fired on 3 of 168 outputs and had sensitivity of 0.00 and 0.19, all
contrasts are descriptive and do not adjudicate H2 or H3.
\end{minipage}
\end{table}

\textbf{Sample selection.} The Probe 2 sample is also selected by outcome-adjacent criteria. The
analysis rests on 42 of 199 pairs, and its dominant source of attrition,
the strict gold-signature requirement, conditions on a property of the
accepted patch that plausibly tracks how distinctive the fix is, so the
surviving pairs need not represent the matched cohort. Selection was
asymmetric earlier in cohort construction: control candidates, unlike
benchmark instances, had to pass a lexical bug-fix screen, and the
concordant sensitivity subset cannot recover controls that were never
collected. The analyzed sample additionally fails the cohort-specific
issue-length balance check, and matching on observable structure does
not rule out unmeasured semantic differences between benchmark and
control tasks.

\begin{table}[htbp]
\caption{Strict-scorer validation against consensus human labels. The
scorer did not pass the validation gate: sensitivity could not be
established for either model, and both DeepSeek-V4-Flash firings were
false positives.}
\label{tab:probe2}
\centering
\footnotesize
\begin{tabular}{@{}l C{1.9cm} C{1.9cm} C{1.7cm} C{2.6cm} C{1.4cm}@{}}
\toprule
\textbf{Model} & \textbf{Human-positive responses}
& \textbf{Detected by strict scorer} & \textbf{Weighted sensitivity}
& \textbf{Exact 95\% CI, unweighted sensitivity}
& \textbf{Precision} \\
\midrule
DeepSeek-V4-Flash & 2/41 & 0/2 & 0.00 & $[0.00, 0.84]$ & 0/2 \\
GPT-4.1           & 3/38 & 1/3 & 0.19 & $[0.01, 0.91]$ & 1/1 \\
\bottomrule
\end{tabular}
\end{table}

\textbf{Manual-validation definitions.} Two authors independently assigned reproduction-validity labels from 0 to 3 while blinded to benchmark-control membership but not model identity or scorer-derived sampling stratum. Labels of 2 or 3 indicated patch-specific implementation evidence, and a response was consensus human-positive when both authors assigned one of these labels. Responses marked undecidable by either author were excluded from sensitivity estimation and reported separately. Sensitivity was the inverse-probability-weighted proportion of consensus human-positive responses detected by the strict scorer. Precision was the proportion of strict-scorer firings on correct-gold comparisons that were consensus human-positive. Cohen's $\kappa$ was computed after binarization at the $\geq 2$ threshold within the enriched sample and is not interpreted as population-level reliability.

\textbf{Manual-review sampling.} For each model, the sampling frame
comprised 84 arms from the final 42 matched pairs. We reviewed all
strict exact hits, all fuzzy-only hits, and the 15 highest-overlap
outputs among the remaining arms. We then sampled 20 outputs without
replacement from the remaining tail using seed 2026. The inverse
inclusion-probability weight was 1 for the first three strata and
$N_{h}/n_{h}$ for the randomly sampled tail.

\begin{table}[htbp]
\caption{Sampling design for the Probe 2 manual scorer validation.}
\label{tab:a2}
\centering
\footnotesize
\begin{tabular}{@{}l l C{1.7cm} C{1.7cm} C{2.1cm} C{1.6cm}@{}}
\toprule
\textbf{Model} & \textbf{Sampling stratum}
& \textbf{Available arms $N_{h}$} & \textbf{Reviewed arms $n_{h}$}
& \textbf{Inclusion probability $\pi_{h}$}
& \textbf{Weight $1/\pi_{h}$} \\
\midrule
\multirow{4}{*}{DeepSeek-V4-Flash}
 & Exact hit    & 2  & 2  & 1.000 & 1.00 \\
 & Fuzzy-only   & 6  & 6  & 1.000 & 1.00 \\
 & High overlap & 15 & 15 & 1.000 & 1.00 \\
 & Random tail  & 61 & 20 & 0.328 & 3.05 \\
\cmidrule(lr){1-6}
\multirow{4}{*}{GPT-4.1}
 & Exact hit    & 1  & 1  & 1.000 & 1.00 \\
 & Fuzzy-only   & 4  & 4  & 1.000 & 1.00 \\
 & High overlap & 15 & 15 & 1.000 & 1.00 \\
 & Random tail  & 64 & 20 & 0.313 & 3.20 \\
\bottomrule
\end{tabular}

\vspace{3pt}
\begin{minipage}{\linewidth}
\footnotesize\raggedright
\textbf{Notes.} Exact-hit outputs were positive under the strict scorer.
Fuzzy-only outputs triggered the fuzzy matcher but not the strict
scorer. The high-overlap stratum comprised the 15 remaining outputs with
the largest sum of token F1 and n-gram overlap. The random-tail stratum
was sampled from all remaining outputs. Sampling was performed
separately for each model. Two responses per model were subsequently
classified as undecidable by at least one author, leaving 41
DeepSeek-V4-Flash and 38 GPT-4.1 responses for sensitivity estimation.
\end{minipage}
\end{table}

\textbf{Robustness checks.} For Probe 1, leave-one-repository-out re-estimation shifted the
four reported repository-balanced gaps by at most 0.068. The
hierarchical Top-3 intervals are reported with the primary results in
Section~\ref{sec:probe1}. In a post-hoc analysis, we restricted Probe 1
to 64 task-type-concordant pairs from all ten repositories. The Top-3
gaps remained positive for both models, but all pair-weighted,
repository-balanced, and hierarchical 95\% intervals included zero
(Table~\ref{tab:a3}). For Probe 2, the 32-pair task-type-concordant
analysis likewise remains descriptive because it does not address the
scorer's limited sensitivity.

\begin{table}[htbp]
\caption{Probe 1 sensitivity analysis restricted to
task-type-concordant pairs.}
\label{tab:a3}
\centering
\footnotesize
\begin{tabular}{@{}l c C{3.3cm} C{3.3cm} C{2.6cm}@{}}
\toprule
\textbf{Model} & \textbf{N pairs}
& \textbf{Pair-weighted gap [95\% CI]}
& \textbf{Repository-balanced gap [95\% CI]}
& \textbf{Hierarchical 95\% CI} \\
\midrule
DeepSeek-V4-Flash & 64 & $+0.063$ $[-0.109, +0.234]$ & $+0.035$ $[-0.206, +0.276]$ & $[-0.244, +0.301]$ \\
GPT-4.1           & 64 & $+0.109$ $[-0.063, +0.281]$ & $+0.128$ $[-0.160, +0.416]$ & $[-0.197, +0.436]$ \\
\bottomrule
\end{tabular}

\vspace{3pt}
\begin{minipage}{\linewidth}
\footnotesize\raggedright
\textbf{Notes.} This post-hoc analysis evaluates Top-3 file-path
accuracy on 64 pairs from ten repositories. A pair is considered
task-type concordant when both arms receive the same rule-based label
(bugfix, maintenance, or other) based only on the issue text shown in
the prompt. Pair-weighted intervals use 10{,}000 paired bootstrap
replicates with seed 7. Repository-balanced intervals are $t$-intervals
computed from ten repository-specific gaps ($df = 9$), while
hierarchical intervals resample repositories first and then pairs within
each repository. The task-type restriction was applied after matching.
All pooled covariates remained below the original
$|\mathrm{SMD}| < 0.20$ criterion (maximum 0.163), although merge timing
(0.236) and deleted lines (0.233) exceeded 0.20 in the smaller 21-pair
HAL-linked subset. We therefore interpret this analysis only as a
post-hoc sensitivity check. The broad task-type categories do not
measure issue clarity or semantic difficulty.
\end{minipage}
\end{table}

\section{Stage I coding reliability and evidence ledger}
\label{app:b}

Both authors independently coded 117 field-level judgments before
viewing the other author's labels. Table~\ref{tab:b1} reports raw
categorical agreement before adjudication; consensus values were not
used to compute these figures. Because the 13 fields use different
response domains and are nested within nine configurations, we report
transparent agreement counts rather than a pooled chance-corrected
coefficient.

\begin{table}[htbp]
\caption{Pre-adjudication agreement for the Stage I susceptibility and
incident card.}
\label{tab:b1}
\centering
\footnotesize
\begin{tabular}{@{}L{4.2cm} c c@{}}
\toprule
\textbf{Agreement unit} & \textbf{Agreements / total}
& \textbf{Agreement} \\
\midrule
Field-level judgments      & 90/117 & 76.9\% \\
Derived T labels           & 9/9    & 100.0\% \\
Derived R labels           & 6/9    & 66.7\% \\
Derived L labels           & 6/9    & 66.7\% \\
All derived T/R/L labels   & 21/27  & 77.8\% \\
Incident channel/status    & 4/9    & 44.4\% \\
\bottomrule
\end{tabular}

\vspace{3pt}
\begin{minipage}{\linewidth}
\footnotesize\raggedright
\textbf{Notes.} Agreement was calculated from the two frozen worksheets
before any consensus values were assigned. The first author subsequently
proposed resolutions for the 27 categorical disagreements, and the
second author reviewed all 27 against the codebook and frozen evidence
packets, accepting 26 and changing one judgment: AssistantBench
\texttt{gold\_agent\_visible} from yes to partial. The final T/R/L
labels were mechanically re-derived from the adjudicated field values.
Scope and rationale were retained for every final judgment. The
anonymized supplementary material contains the two frozen rater snapshots,
field-level confusion tables, proposed and reviewed adjudication files,
final consensus annotations, the adjudication review manifest, and 331 evidence-ID links across all 117 consensus fields.
\end{minipage}
\end{table}

\section{Stage I supplementary audit procedures}
\label{app:c}

Three artifact-based audits supplemented this provenance review. First,
a local inventory mapped task artifacts, prompt construction, retrieval
mechanisms, reference inputs, and evaluator boundaries. Second, for
SWE-bench Verified Mini, we selected 20 tasks using a seed-13 SHA-256
ordering and generated two frozen queries per task: the stable task
identifier and an IDF-maximizing contiguous ten-token span from the
prompt-visible statement. The resulting 40 queries assessed task
discoverability and task-linked gold reachability in the top ten search
results. The focused search audit used OpenAI web search on August 25,
2026, and covered SWE-bench Verified Mini only; evaluation-time
retrieval labels for the other configurations were based on
documentation and code evidence rather than search probing. Third, we
compared locally available benchmark-test and non-test artifacts using
normalized exact matching and five-token shingle Jaccard similarity.
Where execution traces were available, a separate scanner flagged
possible benchmark lookups, gold retrieval, and pipeline leakage.

Search hits, overlap matches, and trace flags were treated only as audit
leads. Confirmation required author review of the original source,
benchmark identity, threat channel, and applicable revision. Search
results were treated as time- and service-dependent observations, with
frozen queries and supporting evidence retained as internal provenance
records. A deterministic validator enforced schema and revision
consistency; missing evidence remained unknown.

\section{Analysis-plan stages, matching cost, and inference settings}
\label{app:d}

\textbf{Stages of methodological decisions.} We distinguish three stages of methodological decisions. The original
analysis plan specified H1--H3, Top-3 as the primary endpoint, the
pair-weighted estimand, the 10{,}000-replicate, seed-7 paired bootstrap,
and the original Probe 2 scoring rule. Throughout the paper, we use
\emph{prespecified} only for choices in this original plan. Before examining model outcomes for the final cohort, we made four outcome-blind amendments to the protocol: we introduced the validation gate, which specified no numerical threshold for adequate scorer sensitivity, introduced the issue-length caliper, set the common-support sample at 199 pairs, and reallocated the repository quotas.
After examining those outcomes, we introduced the strict exact-line scorer as a post-hoc
exploratory analysis and the pair-integrity audit as a post-hoc
data-quality check. We also added the repository-balanced estimand and
its $t$-interval, the hierarchical bootstrap,
leave-one-repository-out analyses, and task-type-concordant analyses as
post-hoc robustness checks. The task-type analyses restrict each probe
to pairs assigned the same rule-based category (bugfix, maintenance, or
other) based on prompt-visible issue text.

\textbf{Control candidates.} Before control collection, we constructed an exclusion registry from
pinned SWE-bench Full, Lite, and Verified revisions to prevent known
benchmark PRs and exact issue or patch duplicates from entering the
control arm. Candidate controls were merged, issue-linked PRs that
passed frozen bug-fix and maintenance screens, changed at least one
production-source file, and added at least one line, leaving 5{,}172
candidates. Matching covariates were computed from the complete GitHub
PR in both arms.

\textbf{Temporal categorization.} For temporal categorization, we
operationalized the GPT-4.1 cutoff as June 1, 2024, 00:00 UTC. A PR was
classified as pre-cutoff if $M \leq C$, post-cutoff if $A > C$, and
straddling if $A \leq C < M$, where $A$ and $M$ denote PR creation and
merge times.

\textbf{Matching cost.} Among feasible pairs, benchmark task $i$ and
control candidate $j$ were assigned the cost
\begin{equation}
d_{ij} = \frac{\Delta_{\mathrm{days}}(M_{i},M_{j})}{365}
 + \left|F_{i} - F_{j}\right|
 + \sum_{k \in K}\frac{\left|x_{ik} - x_{jk}\right|}{s_{k}}
 + \epsilon_{ij},
\label{eq:cost}
\end{equation}
where $M$ is the PR merge date, $F$ is the changed-file count, and $K$
contains issue length, PR-open duration, patch size, additions,
deletions, and source- and test-file counts. PR-open duration, patch
size, additions, and deletions were transformed as $\log(1 + x)$;
$s_{k}$ is the standard deviation computed once over the combined
benchmark and control candidate pool. The deterministic perturbation
$0 \leq \epsilon_{ij} < 10^{-6}$, generated with seed 13, resolves exact
cost ties. Minimum-cost one-to-one assignment was solved separately
within each repository, without control reuse.

\textbf{Post-match balance.} Post-match balance was evaluated on ten
covariates using standardized mean differences (SMDs). We required
$|\mathrm{SMD}| < 0.20$ in the pooled, HAL-linked, and expanded cohorts
and treated 0.10 as a diagnostic target. All covariates satisfied the
0.20 criterion in all three cohorts, and all pooled SMDs were below
0.10, with a maximum absolute SMD of 0.090 for issue length. A
fail-closed validator checked pair eligibility, uniqueness, benchmark
exclusions, balance, and artifact hashes before probe construction.

\textbf{Inference settings.} Both probes used temperature 0,
top-$p = 1$, one completion per task, and no tools, browsing, or
retrieval. Probe 1 used a 128-token output budget. Probe 2 used a fixed
retry ladder up to 2{,}048 tokens. Pairs with persistent truncation were
excluded symmetrically, as described in Appendix~\ref{app:a}.
DeepSeek-V4-Flash ran in non-thinking mode, and GPT-4.1 received no
reasoning parameters.

\section{Hypothesis disposition table}
\label{app:e}

\begin{table}[htbp]
\caption{Disposition of the prespecified hypotheses.}
\label{tab:hyp}
\centering
\footnotesize
\begin{tabular}{@{}L{3.6cm} L{2.7cm} L{6.4cm}@{}}
\toprule
\textbf{Hypothesis} & \textbf{Disposition} & \textbf{Evidence} \\
\midrule
H1: Benchmark tasks exhibit higher file-path accuracy.
& Inconclusive; point estimates were positive, but the primary 95\%
  intervals included zero.
& All gaps were positive; no Top-3 primary interval excluded zero
  (Section~\ref{sec:probe1}). \\
\addlinespace
H2: Benchmark tasks exhibit higher patch-specific line recovery.
& Unresolved: measurement validity insufficient.
& The scorer did not pass the overall validation gate: sufficient
  sensitivity could not be established for either model, and both
  DeepSeek-V4-Flash firings on correct-gold comparisons were
  human-negative. The post-hoc strict-score contrasts therefore remain
  exploratory. \\
\addlinespace
H3: Reproduction is rarer than localization and, conditional on adequate
scorer sensitivity, has an equal or larger benchmark-control gap.
& Unresolved: required sensitivity condition unmet.
& The sensitivity condition could not be established, so neither the
  reproduction-frequency comparison nor the conditional
  benchmark-control gap comparison could be adjudicated. \\
\bottomrule
\end{tabular}
\end{table}

\section{Leakage-screen selection diagnostics}
\label{app:g}

The screen also selects on issue characteristics. Before the
pair-integrity exclusion, the 101 retained benchmark issues averaged 172
words against 328 for the 98 excluded issues, and the retained set was
more HAL-linked (32 of 101 versus 18 of 98). Within-subset covariate
balance (Section~\ref{sec:setup}) therefore does not establish that the
subset represents the cohort: Probe 1 estimates describe a shorter-issue
subpopulation in which explicit path cues were absent from both arms.
The selection shift extended beyond issue length. Compared with the 98
excluded pairs, the 101 screen-passing pairs were less often task-type
concordant under the existing rule-based labels (64.4\% versus 80.6\%)
and contained more changed files on average (3.48 versus 2.92, averaged
across the two arms). Mean pair-level patch-text length was similar
(179.6 versus 174.2 lines, including context and headers), but repository composition differed: Sphinx
accounted for 21.8\% of retained pairs and 8.2\% of excluded pairs,
whereas scikit-learn accounted for 5.9\% and 19.4\%, respectively. These
differences show that the leakage screen changed the analysis population
rather than removing direct cues at random.

\section{Use of large language models and other tools}
\label{app:f}

GPT-4.1 and DeepSeek-V4-Flash are the systems under evaluation in
Stage II; they were queried only as described in
Appendix~\ref{app:d} and were not used to construct the cohort, match
controls, screen for leakage, or score outputs. The focused
search-discoverability audit (Appendix~\ref{app:c}) used the OpenAI
web search service as its search engine on August 25, 2026. No large
language model was used for evidence flagging, data processing or
filtering, visualization, or running experiments; all Stage I
judgments and all Probe 2 reproduction-validity labels were assigned
by the authors.

\newpage
\section*{NeurIPS Paper Checklist}

\begin{enumerate}

\item {\bf Claims}
    \item[] Question: Do the main claims made in the abstract and introduction accurately reflect the paper's contributions and scope?
    \item[] Answer: \answerYes{}
    \item[] Justification: The abstract and Section~\ref{sec:intro} state the three contributions (a susceptibility and incident audit of nine HAL configurations, a matched-control re-evaluation of the file-localization gap, and validation of the reproduction scorer) together with the scope limits: the Probe 1 gap is inconclusive on the primary endpoint, the Probe 2 scorer did not pass its validation gate, and no claim about training-data membership or score inflation is made.

\item {\bf Limitations}
    \item[] Question: Does the paper discuss the limitations of the work performed by the authors?
    \item[] Answer: \answerYes{}
    \item[] Justification: Section~\ref{sec:limitations} discusses scorer validity, outcome-adjacent sample selection, the shorter-issue analysis subset, repository clustering and statistical power, the two-model and ten-repository scope, the absence of confirmed unexposed controls, and the evidence gaps in the Stage I audit; Appendix~\ref{app:a} and Appendix~\ref{app:g} give the underlying selection diagnostics.

\item {\bf Theory assumptions and proofs}
    \item[] Question: For each theoretical result, does the paper provide the full set of assumptions and a complete (and correct) proof?
    \item[] Answer: \answerNA{}
    \item[] Justification: The paper contains no theoretical results; Equations~\ref{eq:pair}, \ref{eq:repo}, and \ref{eq:cost} define estimands and a matching cost, not theorems.

    \item {\bf Experimental result reproducibility}
    \item[] Question: Does the paper fully disclose all the information needed to reproduce the main experimental results of the paper to the extent that it affects the main claims and/or conclusions of the paper (regardless of whether the code and data are provided or not)?
    \item[] Answer: \answerYes{}
    \item[] Justification: Cohort construction, matching, leakage screening, probe definitions, scoring rules, and the analysis plan are specified in Sections~\ref{sec:matched}, \ref{sec:stats}, and \ref{sec:setup} and Appendices~\ref{app:a}, \ref{app:c}, and \ref{app:d}, with pinned benchmark revisions and fixed seeds (13 for sampling and tie-breaking, 7 for the bootstrap, 2026 for manual-review sampling). One caveat is stated in Section~\ref{sec:limitations}: API generation used no fixed seed, so committed model completions are retained but cannot be regenerated bit-for-bit.

\item {\bf Open access to data and code}
    \item[] Question: Does the paper provide open access to the data and code, with sufficient instructions to faithfully reproduce the main experimental results, as described in supplemental material?
    \item[] Answer: \answerYes{}
    \item[] Justification: The anonymized supplementary material contains the Stage I susceptibility card, frozen rater worksheets, adjudication files, and evidence ledger, and the Stage II matched-cohort manifest, prompts, committed model outputs, scoring code, and analysis scripts (Sections~\ref{sec:hal-audit} and~\ref{sec:setup}, Appendix~\ref{app:b}). The evaluated models are commercial API models and are not redistributed.

\item {\bf Experimental setting/details}
    \item[] Question: Does the paper specify all the training and test details (e.g., data splits, hyperparameters, how they were chosen, type of optimizer) necessary to understand the results?
    \item[] Answer: \answerYes{}
    \item[] Justification: No models were trained. The benchmark split, sampling and quota rules, matching calipers and cost function, temporal categorization, balance criteria, decoding settings, output budgets, and truncation handling are given in Section~\ref{sec:setup} and Appendix~\ref{app:d}; scorer construction and manual-review sampling are given in Appendix~\ref{app:a}.

\item {\bf Experiment statistical significance}
    \item[] Question: Does the paper report error bars suitably and correctly defined or other appropriate information about the statistical significance of the experiments?
    \item[] Answer: \answerYes{}
    \item[] Justification: Every Probe 1 gap is reported with a 95\% interval: a prespecified paired percentile bootstrap over matched pairs (10,000 replicates), a post-hoc $t$-interval over repository-specific gaps, and a hierarchical repository-then-pair bootstrap (Section~\ref{sec:stats}, Table~\ref{tab:probe1}). Scorer sensitivity is reported with exact binomial intervals (Table~\ref{tab:probe2}). Intervals are pointwise and unadjusted across model-metric combinations, and the paper states that no family-wise confirmatory claim is made.

\item {\bf Experiments compute resources}
    \item[] Question: For each experiment, does the paper provide sufficient information on the computer resources (type of compute workers, memory, time of execution) needed to reproduce the experiments?
    \item[] Answer: \answerYes{}
    \item[] Justification: No GPU compute was used. All model inference was performed through commercial chat-completion APIs with one completion per task (Appendix~\ref{app:d}): 400 Probe 1 completions and at most 796 Probe 2 completions before truncation retries. Cohort construction, matching, leakage screening, and bootstrap analyses run on a single commodity CPU machine.

\item {\bf Code of ethics}
    \item[] Question: Does the research conducted in the paper conform, in every respect, with the NeurIPS Code of Ethics \url{https://neurips.cc/public/EthicsGuidelines}?
    \item[] Answer: \answerYes{}
    \item[] Justification: The study analyzes public benchmark tasks and public GitHub pull requests, involves no human participants other than the authors, and releases no models. The paper takes care not to present susceptibility labels as accusations of contamination against any benchmark or model (Section~\ref{sec:threat}).

\item {\bf Broader impacts}
    \item[] Question: Does the paper discuss both potential positive societal impacts and negative societal impacts of the work performed?
    \item[] Answer: \answerNo{}
    \item[] Justification: The paper is a methodological audit and has no dedicated broader-impact section. Its main foreseeable effect is on how benchmark scores are interpreted. The principal misuse risk, reading open or partial susceptibility labels as findings of contamination against specific benchmarks or models, is addressed directly in Section~\ref{sec:threat}, the notes to Table~\ref{tab:hal_card}, and Section~\ref{sec:interpretation}.

\item {\bf Safeguards}
    \item[] Question: Does the paper describe safeguards that have been put in place for responsible release of data or models that have a high risk for misuse (e.g., pre-trained language models, image generators, or scraped datasets)?
    \item[] Answer: \answerNA{}
    \item[] Justification: No models are released, and the released data consist of audit records and matched-cohort manifests derived from already-public benchmark tasks and pull requests.

\item {\bf Licenses for existing assets}
    \item[] Question: Are the creators or original owners of assets (e.g., code, data, models), used in the paper, properly credited and are the license and terms of use explicitly mentioned and properly respected?
    \item[] Answer: \answerYes{}
    \item[] Justification: SWE-bench, SWE-bench Verified, HAL, and the prior diagnostic tasks are cited with their original papers, and the benchmark and harness revisions used are pinned (Sections~\ref{sec:hal-audit} and~\ref{sec:setup}). Control pull requests are public GitHub data accessed through the GitHub API under its terms of service and are used for analysis only; the supplementary material references them by repository and PR number rather than redistributing their content.

\item {\bf New assets}
    \item[] Question: Are new assets introduced in the paper well documented and is the documentation provided alongside the assets?
    \item[] Answer: \answerYes{}
    \item[] Justification: The susceptibility and incident card (Section~\ref{sec:threat}, Table~\ref{tab:hal_card}), its codebook and evidence ledger (Appendix~\ref{app:b}), and the matched-cohort construction (Section~\ref{sec:setup}, Appendix~\ref{app:d}) are documented in the paper, and the anonymized supplementary material includes a README describing each file.

\item {\bf Crowdsourcing and research with human subjects}
    \item[] Question: For crowdsourcing experiments and research with human subjects, does the paper include the full text of instructions given to participants and screenshots, if applicable, as well as details about compensation (if any)?
    \item[] Answer: \answerNA{}
    \item[] Justification: No crowdsourcing or external participants were involved; all Stage I coding and Probe 2 manual labels were assigned by the authors (Sections~\ref{sec:hal-audit} and~\ref{sec:matched}).

\item {\bf Institutional review board (IRB) approvals or equivalent for research with human subjects}
    \item[] Question: Does the paper describe potential risks incurred by study participants, whether such risks were disclosed to the subjects, and whether Institutional Review Board (IRB) approvals (or an equivalent approval/review based on the requirements of your country or institution) were obtained?
    \item[] Answer: \answerNA{}
    \item[] Justification: The paper involves no research with human subjects.

\item {\bf Declaration of LLM usage}
    \item[] Question: Does the paper describe the usage of LLMs if it is an important, original, or non-standard component of the core methods in this research? Note that if the LLM is used only for writing, editing, or formatting purposes and does \emph{not} impact the core methodology, scientific rigor, or originality of the research, declaration is not required.
    \item[] Answer: \answerYes{}
    \item[] Justification: GPT-4.1 and DeepSeek-V4-Flash are the objects of evaluation, not components of the audit method; their use is specified in Section~\ref{sec:setup} and Appendix~\ref{app:d}. Appendix~\ref{app:f} states that no LLM was used for evidence flagging, data processing, scoring, or manual labeling, and identifies the web search service used in the discoverability audit.

\end{enumerate}

\end{document}